\documentclass[10pt,twocolumn,letterpaper]{article}

\usepackage{cvpr}
\usepackage{times}
\usepackage{epsfig}
\usepackage{graphicx}
\usepackage{amsmath}
\usepackage{amssymb}
\usepackage{float}
\usepackage[english]{babel}
\usepackage{blindtext}
\usepackage{gensymb}
\usepackage{array}
\usepackage{url}
\usepackage{subfigure}
\usepackage{makecell}
\newcolumntype{C}[1]{>{\centering\let\newline\\\arraybackslash\hspace{0pt}}m{#1}}

\cvprfinalcopy 

\def\cvprPaperID{****} 

\begin{document}

\title{Optical Flow Estimation using a Spatial Pyramid Network}

\author{Anurag Ranjan \quad \quad Michael J. Black\hspace{0.1in}\\ 
    Max Planck Institute for Intelligent Systems, T\"{u}bingen, Germany\\
    {\tt\small \{anurag.ranjan, black\}@tuebingen.mpg.de} 
       }

\maketitle

\begin{abstract}
We learn to compute optical flow by combining a classical spatial-pyramid formulation with deep learning.
This estimates large motions in a coarse-to-fine approach by
warping one image of a pair at each pyramid level by the current flow estimate and computing an update to the flow.
Instead of the standard minimization of an objective function at each
pyramid level, we train one deep network per level to compute the flow update.
Unlike the recent FlowNet approach, the networks do not need to deal with large motions; these are dealt with by the pyramid.  
This has several advantages. 
First, our Spatial Pyramid Network (SPyNet) is much simpler and  
96\% smaller than FlowNet in terms of model parameters.
This makes it more efficient and appropriate for embedded applications.
Second, since the flow at each pyramid
level is small ($< 1$ pixel), a convolutional approach applied to pairs of warped images is appropriate.
Third, unlike FlowNet, the learned convolution filters appear similar to classical spatio-temporal filters, 
giving insight into the method and how to improve it.
Our results are more accurate than FlowNet on most standard benchmarks, suggesting a new direction of combining classical flow methods with deep learning. 
\end{abstract}
\section{Introduction}
Recent years have seen significant progress on the problem of accurately estimating optical flow, as evidenced by improving performance on increasingly challenging benchmarks.
Despite this, most flow methods are derived from a ``classical formulation'' that makes a variety of assumptions about the image, from brightness constancy to spatial smoothness.
These assumptions are only coarse approximations to reality and this likely limits performance.
The recent history of the field has focused on improving these assumptions or making them more robust to violations \cite{black1993framework}.
This has led to steady but incremental progress.

An alternative approach abandons the classical formulation altogether and starts over using recent neural network architectures.  
Such an approach takes a pair (or sequence) of images and learns to directly compute flow from them.
Ideally such a network would learn to solve the correspondence problem (short and long range), learn filters relevant to the problem, learn what is constant in the sequence, and learn about the spatial structure of the flow and how it relates to the image structure.
The first attempts are promising but are not yet as accurate as the classical methods.

{\bf Goal.}
We argue that there is an alternative approach that combines the best of both approaches.
Decades of research on flow has produced well engineered systems and principles that are effective.
But there are places where these methods make assumptions that limit their performance.
Consequently, here we apply machine learning to address the weak points, while keeping the engineered architecture, with the goal of 1) improving performance over existing neural networks and the classical methods upon which our work is based;
2) achieving real-time flow estimates with accuracy better than the much slower classical methods; and 3) reducing memory requirements to make flow more practical for embedded, robotic, and mobile applications.

{\bf Problem.} 
The key problem with recent methods for learning flow \cite{dosovitskiy2015flownet} 
is that they typically take two frames, stack them together, and apply a convolutional network architecture.
When the motions between frames are larger than one (or a few) pixels, spatio-temporal convolutional filters will not obtain meaningful responses.
Said another way, if a convolutional window in one image does not overlap with related image pixels at the next time instant, no meaningful temporal filter can be learned.

There are two problems that need to be solved.
One is to solve for long-range correlations while the other is to solve for detailed, sub-pixel, optical flow and precise motion boundaries.
FlowNet \cite{dosovitskiy2015flownet} attempts to learn both of these at once.
In contrast, we tackle the latter using deep learning and rely on existing methods to solve the former.

{\bf Approach.}
To deal with large motions we adopt a traditional coarse-to-fine approach using a spatial pyramid\footnote{This, of course, has well-known limitations, which we discuss later.}.
At that top level of the pyramid, the hope is that the motions between frames are smaller than a few pixels and that, consequently, the convolutional filters can learn meaningful temporal structure.
At each level of the pyramid we solve for the flow using a convolutional network and
up-sample the flow to the next pyramid level.
As is standard, with classical formulations \cite{sun2014quantitative},
we warp one image towards the other using the current flow, and repeat this process at each pyramid level.
Instead of minimizing a classical objective function at each level, we learn a convolutional network to predict the {\em flow increment} at that level.
We train the network from coarse to fine to learn the flow correction at each level and add this to the flow output of the network above.
The idea is that the displacements are then always less than a few pixels at each pyramid level.

We call the method {\em SPyNet}, for Spatial Pyramid Network, and train it using the same Flying Chairs data as FlowNet \cite{dosovitskiy2015flownet}.
We report similar performance as FlowNet on Flying Chairs and Sintel \cite{Butler:ECCV:2012} but are  
significantly more accurate than FlowNet on Middlebury \cite{baker2011database} and KITTI \cite{Geiger2012CVPR} after fine tuning.
The total size of SPyNet is 96\% smaller than FlowNet, meaning that it runs faster, and uses much less memory.
%
The expensive iterative propagation of classical methods is replaced by the non-iterative computation of the neural network.

We do not claim to solve the full optical flow problem with SPyNet -- we address the same problem as traditional approaches and inherit some of their limitations.  
For example, it is well known that large motions of small or thin objects are difficult to capture with a pyramid representation.  
We see the large motion problem as separate, requiring different solutions.
Rather, what we show is that the traditional problem can be reformulated, portions of it can be learned, and performance improves in many scenarios.

Additionally, because our approach connects past methods with new tools, it provides insights into how to move forward.
In particular, we find that SPyNet learns spatio-temporal convolutional filters that resemble traditional spatio-temporal derivative or Gabor filters \cite{adelson1985spatiotemporal,heeger1987model}.
The learned filters resemble biological models of motion processing filters in MT and V1 
\cite{Simoncelli1998}. 
This is in contrast to the highly random-looking filters learned by FlowNet.
This suggests that it is timely to reexamine older spatio-temporal filtering approaches with new tools.

In summary our contributions are:
1) the combination of traditional coarse-to-fine pyramid methods with deep learning for optical flow estimation;
2) a new SPyNet model that is 96\% smaller and faster than FlowNet;
3) SPyNet achieves comparable or lower error than FlowNet on standard benchmarks -- Sintel, KITTI and Middlebury;
4) the learned spatio-temporal filters provide insight about what filters are needed for flow estimation;
5) the trained network and related code are publicly available for research \footnote{\url{https://github.com/anuragranj/spynet}}.

\section{Related Work}
Our formulation effectively combines ideas from ``classical'' optical flow and recent deep learning methods.
Our review focuses on the work most relevant to this.

{\bf Spatial pyramids and optical flow.}
The classical formulation of the optical flow problem dates to Horn and Schunck \cite{horn1981determining} and involves optimizing the sum
of a data term based on brightness constancy and a spatial smoothness term.
The classical methods typically suffer from the fact that they make very approximate assumptions about the image brightness change and the spatial structure of the flow.
Many methods focus on improving robustness by changing the assumptions.
A full review would effectively cover the history of the field; for this we refer the reader to \cite{sun2014quantitative}. 
The key advantage of learning to compute flow, as we do here, is that we do not hand craft changes in these assumptions.
Rather, the variation in image brightness and spatial smoothness are embodied in the learned network.


The idea of using a spatial pyramid has a similarly long history dating to \cite{burt-adelson} with its first use in the classical flow formulation appearing in
\cite{glazer-thesis}. 
Typically Gaussian or Laplacian pyramids are used for flow estimation with the primary motivation to deal with large motions.
These methods are well known to have problems when small objects move quickly.
Brox et al.~\cite{brox2009large} incorporate long range matching into the traditional optical flow objective function.
This approach of combining image matching to capture large motions, with a variational \cite{epicflow} or discrete optimization \cite{guney2016ACCV} for fine motions, can produce accurate results.




Of course spatial pyramids are widely used in other areas of computer vision and have recently been used in deep neural networks \cite{denton2015deep} to learn generative image models.

{\bf Spatio-temporal filters.}
Burt and Adelson \cite{adelson1985spatiotemporal} lay out the theory of spatio-temporal models for motion estimation  and 
Heeger \cite{heeger1987model} provides a computational embodiment.
While inspired by human perception, such methods did not perform well at the time \cite{barron94-ijcv}.


Various methods have shown that spatio-temporal filters emerge from learning, for example using independent component analysis \cite{vanHateren:1998}, sparseness \cite{olshausen2003learning}, and multi-layer models \cite{cadieu2008learning}.
Memisevic and Hinton learn simple spatial transformations with a restricted Boltzmann machine \cite{memisevic2010learning}, finding a variety of filters.
Taylor et al.~\cite{taylor2010convolutional} use synthetic data to learn ``flow like'' features using a restricted Boltzmann machine but do not evaluate flow accuracy.

Dosovitskiy et al.~\cite{dosovitskiy2015flownet} learn spatio-temporal filters for flow estimation using a deep network, yet these filters do not resemble classical filters inspired by neuroscience.
By using a pyramid approach, here we learn filters that are visually similar to classical spatio-temporal filters, yet because they are learned from data, produce good flow estimates.




{\bf Learning to model and compute flow.}
Possibly the first attempt to learn a model to estimate optical flow is the work of Freeman et al.~\cite{Freeman2000} using an MRF.
They consider a simple synthetic world of uniform moving blobs with ground truth flow.
The training data was not realistic and they did not apply the method to real image sequences. 

Roth and Black \cite{roth2009fields} learn a field-of-experts (FoE) model to capture the spatial statistics of optical flow.
The FoE can be viewed as a (shallow) convolutional neural network.
The model is trained using flow fields generated from laser scans of real scenes and natural camera motions.
They have no images of the scenes (only their flow) and consequently the method only learns the spatial component.

Sun et al.~\cite{roth2008learning} describe the first fully learned model that can be considered a (shallow) convolutional neural network.  
They formulate a classical flow problem with a data term and a spatial term.  The spatial term uses the FoE model from \cite{roth2009fields}, while the data term replaces traditional derivative filters with a set of learned convolutional image filters. 
With limited training data and a small set of filters, it did not fully show the full promise of learning flow.


Wulff and Black \cite{wulff2015efficient} learn the spatial statistics of optical flow by a
applying robust PCA \cite{Hauberg:PAMI:2015} to real (noisy) optical flow computed from natural movies.
While this produces a global flow basis and overly smooth flow, 
they use the model to compute reasonable flow relatively quickly.






{\bf Deep Learning.} 
The above learning methods suffer from limited training data and the use of shallow models.
In contrast, deep convolutional neural networks have emerged as a powerful class of models for solving recognition \cite{he2015deep,szegedy2015going} and dense estimation \cite{chen2014semantic,long2015fully} problems. 


FlowNet \cite{dosovitskiy2015flownet} represents the first deep convolutional architecture for flow estimation that is trained end-to-end. 
The network shows promising results, despite being trained on an artificial dataset of chairs flying over randomly selected images.
Despite promising results, the method lags behind the state of the art in terms of accuracy \cite{dosovitskiy2015flownet}.  
Deep matching methods \cite{guney2016ACCV,epicflow,weinzaepfel2013deepflow,thewlis2016fully} do not fully solve the problem, since they resort to classical methods to compute the final flow field.
It remains an open question as to which architectures are most appropriate for the problem and how best to train these.



Tran et al.~\cite{Tran:End2End:2016}, use a traditional flow method to create ``semi-truth'' training data for a 3D convolutional network. 
The performance is below the state of the art and the method is not tested on the standard benchmarks.
There have also been several attempts at estimating optical flow using unsupervised learning \cite{ahmadi2016unsupervised,yu2016back}. 
However these methods have lower accuracy on standard benchmarks.

{\bf Fast flow.} Several recent methods attempt to balance speed and accuracy, with the goal of real-time processing and reasonable (though not top) accuracy.
GPU-flow \cite{Werlberger2009:GPUflow} began this trend but several methods now outperform it.
PCA-Flow \cite{wulff2015efficient} runs on a CPU, is slower than frame rate, and produces overly smooth flow fields.
EPPM \cite{bao2014tipeppm} achieves similar, middle-of-the-pack, performance on Sintel (test), with similar speed on a GPU.
Most recently DIS-Fast \cite{Kroeger:ECCV:2016} is a GPU method that is significantly faster than previous methods but is also significantly less accurate.

Our method is also significantly faster than the best previous CNN flow method (FlowNet), which reports a runtime of 80ms/frame for FlowNetS. 
The key to our speed is to create a small neural network that fits entirely on the GPU. 
Additionally all our pyramid operations are implemented on the GPU.

Size is an important issue that has not attracted as much attention as speed.
For optical flow to exist on embedded processors, aerial vehicles, phones, etc., the algorithm needs a small memory footprint.
Our network is 96\% smaller than FlowNetS and uses only 9.7 MB for the model parameters, making it easily small enough to fit on a mobile phone GPU.

\section{Spatial Pyramid Network}   
Our approach uses the coarse-to-fine spatial pyramid structure of \cite{denton2015deep} to learn residual flow at each pyramid level. 
Here we describe the network and training procedure. 

\begin{figure*}
\centerline{
\includegraphics[width=\linewidth]{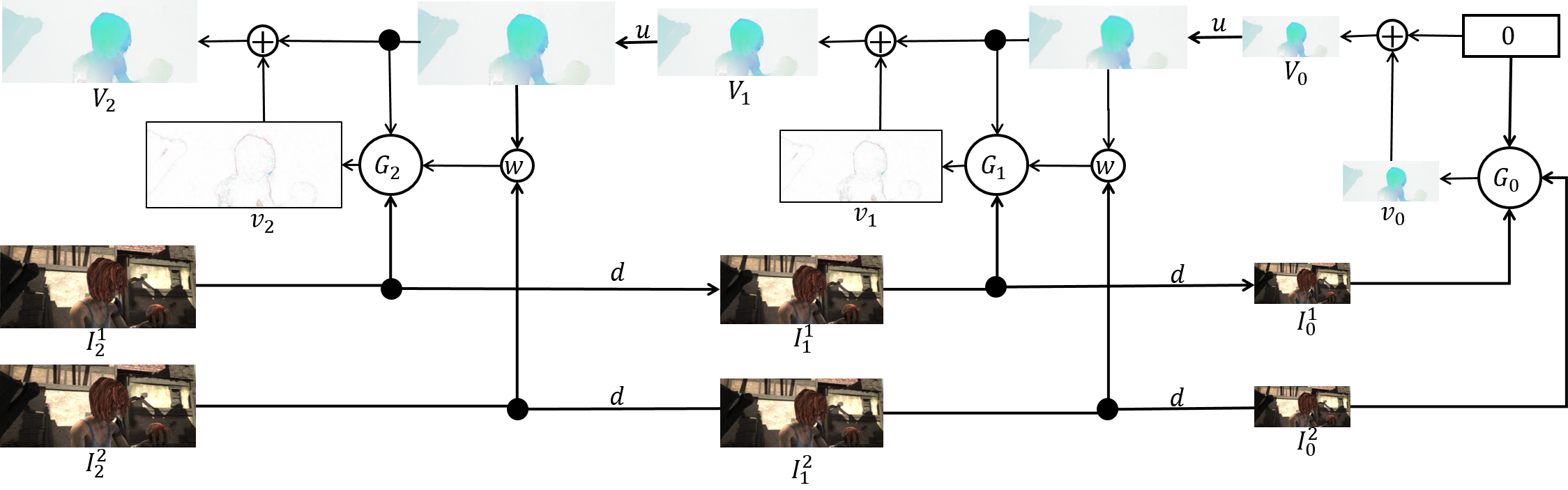}
}
\caption{Inference in a 3-Level Pyramid Network \cite{denton2015deep}: The network $G_0$ computes the residual flow $v_0$ at the highest level of the pyramid (smallest image) using the low resolution images $\{I_0^1, I_0^2\}$. At each pyramid level, the network $G_k$ computes a residual flow $v_k$ which propagates to each of the next lower levels of the pyramid in turn, to finally obtain the flow $V_2$ at the highest resolution.}
\label{fig:infer}
\end{figure*}

\subsection{Spatial Sampling}
Let $d(\cdot)$ be the downsampling function that decimates an $m \times n$ image $I$ to the corresponding image $d(I)$ of size $m/2 \times n/2$. Let $u(\cdot)$ be the reverse operation that upsamples images. These operators are also used for downsampling and upsampling the horizontal and vertical components of the optical flow field, $V$. 
We also define a warping operator $w(I,V)$ that warps the image, $I$ according to the flow field, $V$, using  bi-linear interpolation.

\subsection{Inference}
Let $\{G_0, ... ,G_K\}$ denote a  set of trained convolutional neural network (convnet) models,
each of which computes residual flow, $v_k$
\begin{align}
v_k &= G_k(I_k^1, w(I_k^2,u(V_{k-1}) ), u(V_{k-1}))
\end{align}
at the $k$-th pyramid level. The convnet $G_k$ computes the residual flow $v_k$ using the upsampled flow from the previous pyramid level, ${V_{k-1}}$, and the frames $\{ I^1_k, I^2_k \}$ at level $k$. 
The second frame $I^2_k$ is warped using the flow as $w(I^2_k, u({V_{k-1}}))$ before feeding it to the convnet $G_k$. The flow, $V_k$ at the $k$-th pyramid level is then 
\begin{align}
\label{eq:flowprop}
V_k &= u(V_{k-1}) + v_k .
\end{align}
As shown in Fig.~\ref{fig:infer}, we start with downsampled images $\{ I_0^1, I_0^2\}$ and an initial flow estimate that is zero everywhere to compute the residual flow $v_0 = V_0$ at the top of the pyramid. We upsample the resulting flow, $u(V_0)$, and pass it to the network $G_1$ along with $\{I_1^1, w(I_1^2, u(V_0)) \}$ to compute the residual flow $v_1$. At each pyramid level, we compute the flow $V_k$ using Equation (\ref{eq:flowprop}). The flow $V_k$ is similarly propagated to higher resolution layers of the pyramid until we obtain the flow $V_K$ at full resolution. 
Figure \ref{fig:infer} shows the working of our approach using a 3-level pyramid. In experiments, we use a 5-level pyramid ($K=4$).

\begin{figure}[t]
\centerline{   
\includegraphics[width=\linewidth]{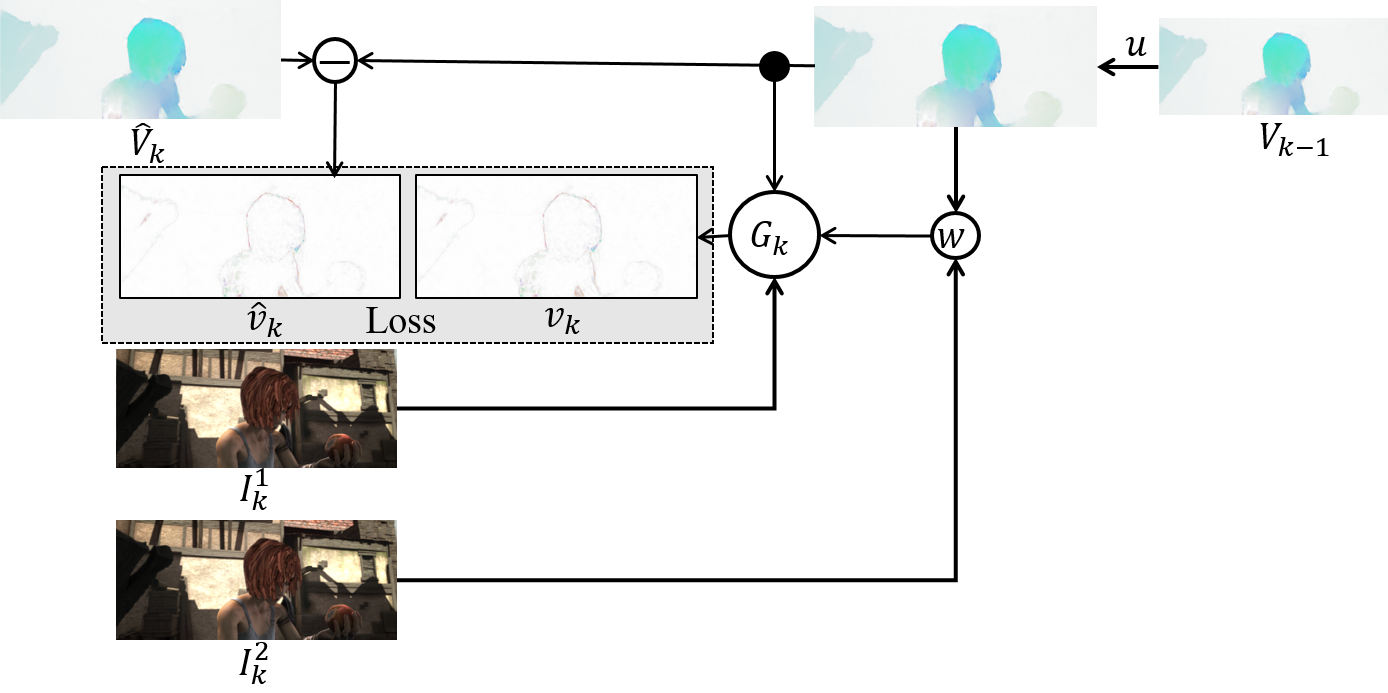}
}
   \caption{Training network $G_k$ requires trained models $\{G_0 ... G_{k-1}\}$ to obtain the initial flow $u(V_{k-1})$. We obtain ground truth residual flows $\hat{v}_k$ by subtracting downsampled ground truth flow $\hat{V}_k$ and $u(V_{k-1})$ to train the network $G_k$ using the EPE loss.}
\label{fig:Train}
\end{figure}

\subsection{Training and Network Architecture}
We train each of the convnets $\{G_0,...,G_K\}$ independently and sequentially to compute the residual flow $v_k$ given the inputs $\{I_k^1, w(I_k^2,u(V_{k-1}) ), u(V_{k-1})\}$. We compute target residual flows $\hat{v}_k$ as a difference of target flow $V_k$ at the $k$-th pyramid level and the upsampled flow, $u(V_{k-1})$ obtained from the trained convnet of the previous level
\begin{align}
\hat{v}_k = \hat{V}_k - u(V_{k-1}) .
\end{align}
As shown in Fig.~\ref{fig:Train}, we train each of the networks, $G_k$, to minimize the average End Point Error (EPE) loss on the residual flow $v_k$.

Each level in the pyramid has a simplified task relative to the full optical flow estimation problem; it only has to estimate a small-motion update to an existing flow field.
Consequently each network can be simple.
Here, each $G_k$ has 5 convolutional layers, which we found gave the best combination of accuracy, size, and speed.
We train five convnets $\{G_0, ..., G_4 \}$ at different resolutions of the Flying Chairs dataset. The network $G_0$ is trained with 24x32 images. We double the resolution at each lower level and finally train the convnet, $G_4$ with a resolution of 384x512.

Each convolutional layer is followed by a Rectified Linear Unit (ReLU), except the last one. 
We use a 7x7 convolutional kernel for each of the layers and found these work better than smaller filters.
The number of feature maps in each convnet, $G_k$ are \{32, 64, 32, 16, 2\}. The image $I^1_k$ and the warped image $w(I^2_k, u(V_{k-1}))$ have 3 channels each (RGB). The upsampled flow $u(V_{k-1})$ is 2 channel (horizontal and vertical).  
We stack image frames together with upsampled flow to form an 8 channel input to each $G_k$. The output is 2 channel flow corresponding to velocity in $x$ and $y$ directions. 


We train five networks $\{G_0, ..., G_4 \}$ such that each network $G_k$ uses the previous network $G_{k-1}$ as initialization. The networks are trained using Adam \cite{kingma2014adam} optimization with $\beta_1 = 0.9$ and $\beta_2 = 0.999$. We use a batch size of 32 across all networks with 4000 iterations per epoch. We use a learning rate of 1e-4 for the first 60 epochs and decrease it to 1e-5 until the networks converge. We use Torch7\footnote{\url{http://torch.ch/}} as our deep learning framework. We use the Flying Chairs \cite{dosovitskiy2015flownet} dataset and the MPI Sintel \cite{Butler:ECCV:2012} for training our network. 
All our networks are trained on a single Nvidia K80 GPU.

We include various types of data augmentation during training. We randomly scale images by a factor of $[1, 2]$ and apply rotations at random within $[-17^{\circ}, 17^{\circ}]$. We then apply a random crop to match the resolution of the convnet, $G_k$ being trained. We include additive white Gaussian noise sampled uniformly from $\mathcal{N}(0, 0.1)$. We apply color jitter with additive brightness, contrast and saturation sampled from a Gaussian, $\mathcal{N}(0, 0.4)$. We finally normalize the images using a mean and standard deviation computed from a large corpus of ImageNet \cite{ILSVRC15} data in \cite{he2015deep}.

\section{Experiments}
\begin{figure}[t]
\begin{center}
\includegraphics[width=\linewidth]{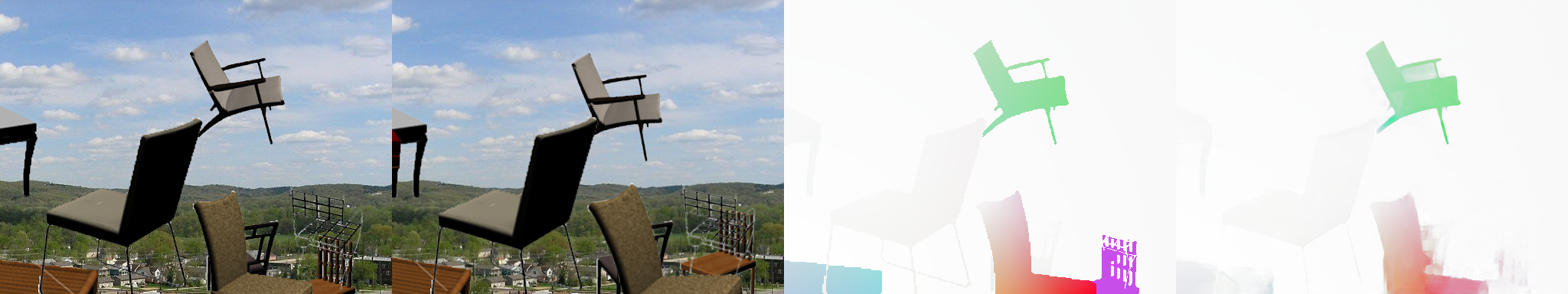}
\includegraphics[width=\linewidth]{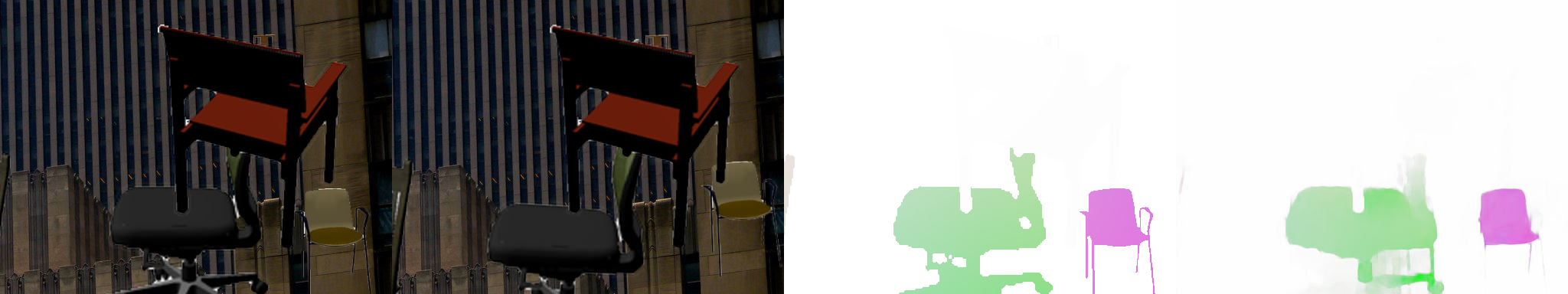}
\includegraphics[width=\linewidth]{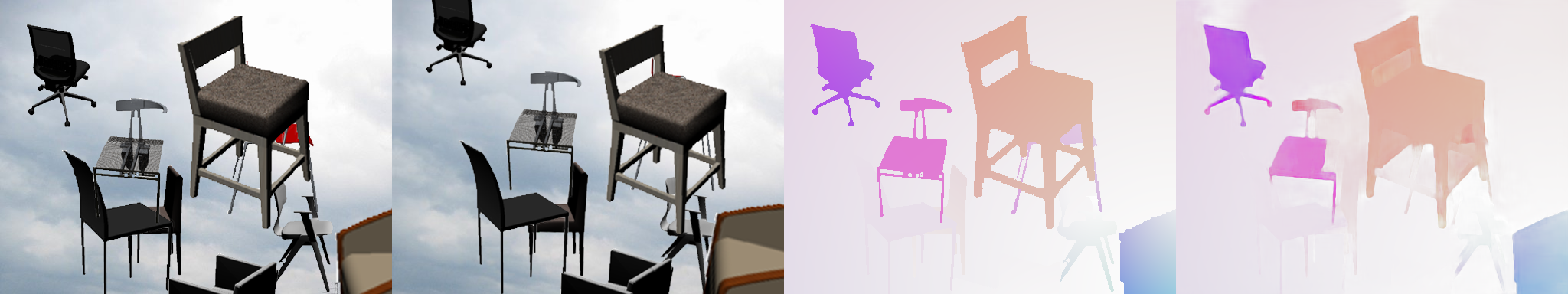}
\includegraphics[width=\linewidth]{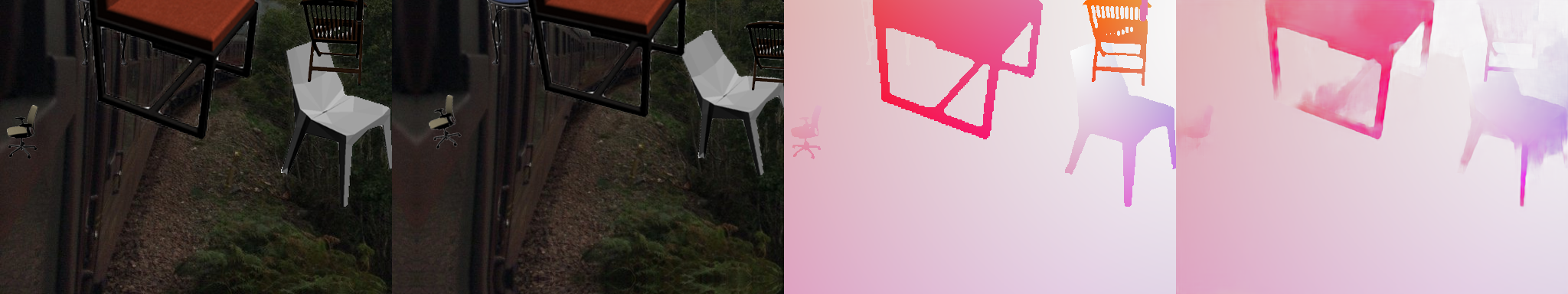}
\begin{tabular}{C{0.08\textwidth}C{0.09\textwidth}C{0.14\textwidth}C{0.07\textwidth}}
Frame 1 & Frame 2 & Ground Truth & SPyNet \\
 \end{tabular}
\end{center}
\vspace{-0.1in}
   \caption{Visualization of optical flow estimates using our model (SPyNet) and the corresponding ground truth flow fields on the Flying Chairs dataset.}
\label{fig:chairsresults}
\end{figure}

\begin{figure*}
\begin{center}
\includegraphics[width=\linewidth]{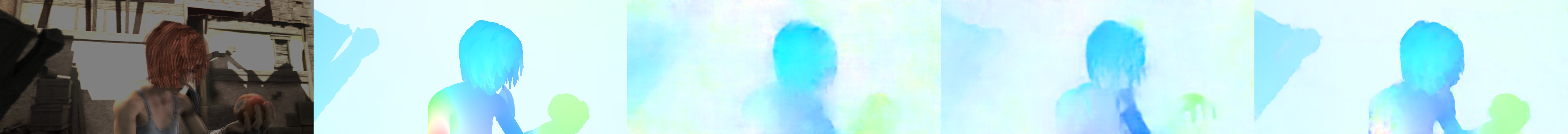}
\includegraphics[width=\linewidth]{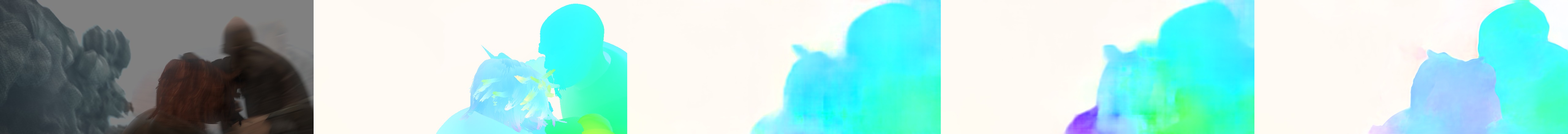}
\includegraphics[width=\linewidth]{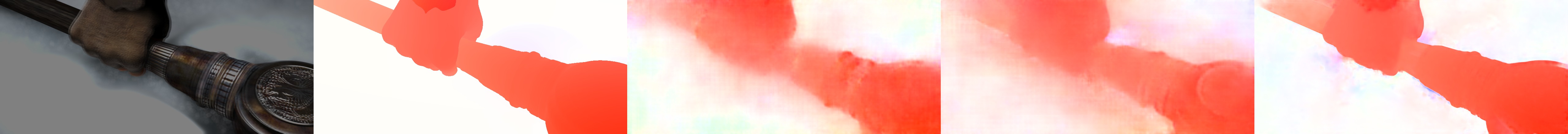}
\includegraphics[width=\linewidth]{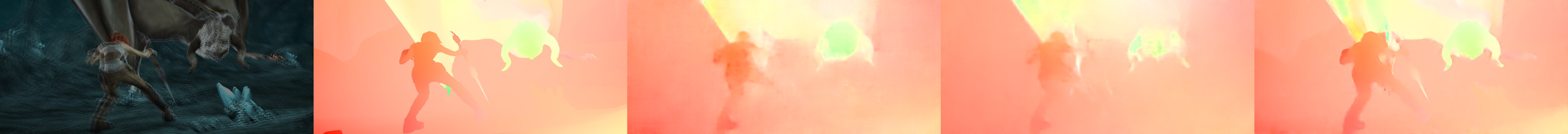}
\includegraphics[width=\linewidth]{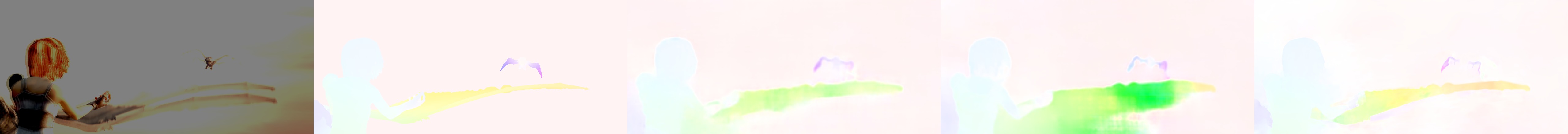}
\includegraphics[width=\linewidth]{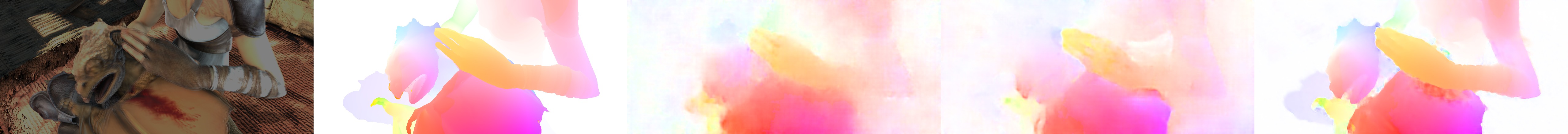}
\includegraphics[width=\linewidth]{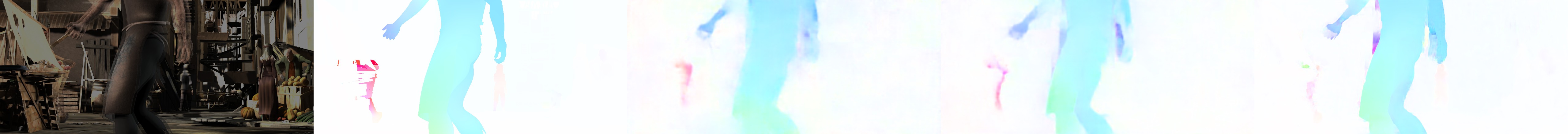}
\includegraphics[width=\linewidth]{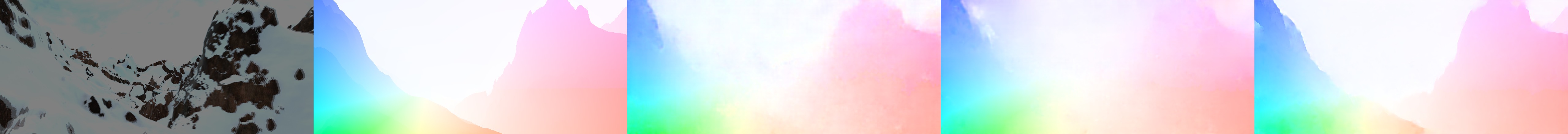}
\includegraphics[width=\linewidth]{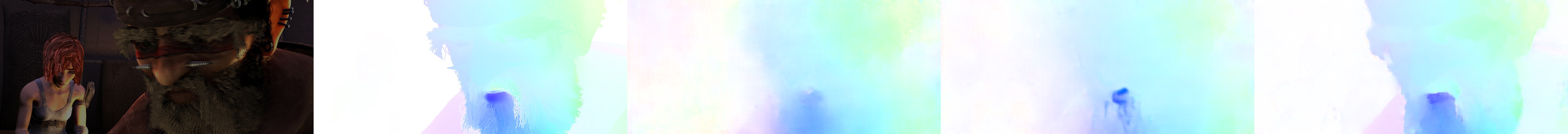}
\includegraphics[width=\linewidth]{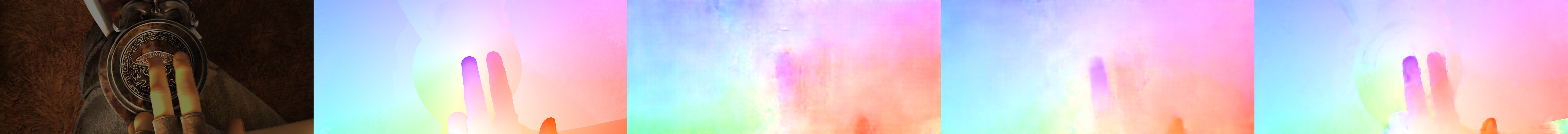}
\begin{tabular}{C{0.2\textwidth}C{0.16\textwidth}C{0.16\textwidth}C{0.16\textwidth}C{0.2\textwidth}}
Frames & Ground Truth & FlowNetS & FlowNetC & SPyNet \\
 \end{tabular}
\end{center}
\vspace{-0.1in}
\caption{Visual comparison of optical flow estimates using our  SPyNet model with FlowNet on the MPI Sintel dataset. The top five rows are from the  Sintel Final set and the bottom five row are from the Sintel Clean set.
SPyNet performs particularly well when the motions are relatively small.
}
\label{fig:sintelresults}
\end{figure*}

\begin{table*}
\begin{center}
\begin{tabular}{lcccccccccc}
\Xhline{4\arrayrulewidth}
Method      & \multicolumn{2}{c}{\underline{Sintel Clean}} & \multicolumn{2}{c}{\underline{Sintel Final}} & \multicolumn{2}{c}{\underline{  \quad  KITTI \quad    }} & \multicolumn{2}{c}{\underline{Middlebury}} & \underline{Flying Chairs} & \multicolumn{1}{l}{Time (s)} \\
            & train            & test           & train            & test           & train        & test        & train           & test          & test          & \multicolumn{1}{l}{}   \\ \Xhline{4\arrayrulewidth}
Classic+NLP    & 4.13             & 6.73           & 5.90             & 8.29           & -         & -           & 0.22            & 0.32             & 3.93          & 102                         \\ \hline
FlowNetS    & 4.50             & 7.42           & \textbf{5.45}             & \textbf{8.43}           & \textbf{8.26}         & -           & 1.09            & -             & 2.71          & 0.080                         \\
FlowNetC    & 4.31             & 7.28           & 5.87             & 8.81           & 9.35         & -           & 1.15            & -             & \textbf{2.19}          & 0.150                         \\
SPyNet      & \textbf{4.12}             & \textbf{6.69}           & 5.57             & \textbf{8.43}           & 9.12         & -             & \textbf{0.33}            & \textbf{0.58}          & 2.63          & \textbf{0.069}                         \\ \hline
FlowNetS+ft & 3.66             & 6.96           & 4.44             & \textbf{7.76}           & 7.52         & 9.1         & 0.98            & -             & 3.04          & 0.080                         \\
FlowNetC+ft & 3.78             & 6.85           & 5.28             & 8.51           & 8.79         & -           & 0.93            &               & \textbf{2.27}          & 0.150                         \\
SPyNet+ft   & \textbf{3.17}             & \textbf{6.64}           & \textbf{4.32 }            & 8.36           & \textbf{4.13}         & \textbf{4.7}         & \textbf{0.33}            & \textbf{0.58}           & 3.07             & \textbf{0.069}                         \\  \Xhline{4\arrayrulewidth}
\end{tabular}
\end{center}
\caption{Average end point errors (EPE). Results are divided into methods trained with (+ft) and without fine tuning.  Bold font indicates the most accurate results among the convnet methods. All run times are measured on Flying Chairs and exclude image loading time.}
\label{tab:results}
\end{table*}

\begin{table*}
\begin{center}
\begin{tabular}{lcccccccccccc}
\Xhline{4\arrayrulewidth}
\multicolumn{1}{l}{Method} & \multicolumn{6}{c}{\underline{\qquad \qquad \qquad \quad Sintel Final \quad \qquad \qquad \qquad}} & \multicolumn{6}{c}{\underline{\qquad \qquad \qquad \quad Sintel Clean \quad \qquad \qquad \qquad}} \\ 
 & $d_{0\text{-}10}$ & $d_{10\text{-}60}$ & $d_{60\text{-}140}$ & $s_{0\text{-}10}$ & $s_{10\text{-}40}$ & $s_{40+}$ & $d_{0\text{-}10}$ & $d_{10\text{-}60}$ & $d_{60\text{-}140}$ & $s_{0\text{-}10}$ & $s_{10\text{-}40}$ & $s_{40+}$\\ \hline
FlowNetS+ft & 7.25 & 4.61 &\textbf{ 2.99} & 1.87 & 5.83 & 43.24 & 5.99 & 3.56 & 2.19 & 1.42 & 3.81 & 40.10 \\ 
FlowNetC+ft & 7.19 & 4.62 & 3.30 & 2.30 & 6.17 & \textbf{40.78} & 5.57 & 3.18 & 1.99 & 1.62 & 3.97 & \textbf{33.37 }\\ 
SpyNet+ft & \textbf{6.69} & \textbf{4.37} & 3.29 &\textbf{ 1.39} & \textbf{5.53 }& 49.71 & \textbf{5.50} & \textbf{3.12} & \textbf{1.71 }& \textbf{0.83} & \textbf{3.34 }& 43.44 \\ \Xhline{4\arrayrulewidth}
\end{tabular}
\end{center}
\caption{Comparison of FlowNet and SpyNet on the Sintel benchmark for different velocities, $s$, and distances, $d$, from motion boundaries.}
\label{tab:sintel}
\end{table*}

We evaluate our performance on standard optical flow benchmarks and compare with FlowNet \cite{dosovitskiy2015flownet}
and Classic+NLP \cite{sun2014quantitative}, a traditional pyramid-based method.
We compare performance using average end point errors in Table \ref{tab:results}.
We evaluate on all the standard benchmarks and find that 
SPyNet is the most accurate overall, with and without fine tuning (details below).
Additionally SPyNet is faster than all other methods. 

Note that the FlowNet results reported on the MPI-Sintel website are for a version that applies variational refinement (``+v'') to the convnet results.
Here we are not interested in the variational component and only compare the results of the convnet output.

\paragraph{Flying Chairs.}
Once the convnets $G_k$ are trained on Flying Chairs, we fine tune the network on the same dataset but without any data augmentation at a learning rate of 1e-6. 
We see an improvement of EPE by 0.14 on the test set. 
Our model achieves better performance than FlowNetS \cite{dosovitskiy2015flownet} on the Flying Chairs dataset, however FlowNetC \cite{dosovitskiy2015flownet} performs better than ours. 
We show the qualitative results on Flying Chairs dataset in Fig.~\ref{fig:chairsresults} and compare the performance in Table \ref{tab:results}.

\paragraph{MPI-Sintel.}
The resolution of Sintel images is 436x1024. 
To use SPyNet, we scale the images to 448x1024, and use 6 pyramid levels to compute the optical flow. The networks used on each pyramid level are $\{G_0, G_1, G_2, G_3, G_4, G_4\}$. We repeat the network $G_4$ at the sixth level of pyramid for experiments on Sintel. 
Because Sintel has extremely large motions, we found that this gives better performance than using just five levels.

We evaluate the performance of our model on MPI-Sintel \cite{Butler:ECCV:2012} in two ways.
First, we directly use the model trained on Flying Chairs dataset and evaluate our performance on both the training and the test sets. 
Second, we extract a validation set from the Sintel training set, using  the same partition as \cite{dosovitskiy2015flownet}. 
We fine tune our model independently on the Sintel Clean and Sintel Final split, and evaluate the EPE. 
The fine-tuned models are listed as ``+ft'' in Table \ref{tab:results}.
We show the qualitative results on MPI-Sintel in Fig.~\ref{fig:sintelresults}. 

Table \ref{tab:sintel} compares our fine-tuned model with FlowNet \cite{dosovitskiy2015flownet} for different velocities and distances from motion boundaries. 
We observe that SPyNet is more accurate than FlowNet for all velocity ranges  except the largest displacements (over 40 pixels/frame). 
SPyNet is also more accurate than FlowNet close to motion boundaries, which is important for many problems.  

\paragraph{KITTI and Middlebury.}
We evaluate KITTI \cite{Geiger2012CVPR} scenes using the base model SPyNet trained on Flying Chairs. 
We then fine-tune the model on Driving and Monkaa scenes from \cite{sceneflowdataset} and evaluate the fine-tuned model SPyNet+ft. 
Fine tuning results in a significant improvement in accuracy by about 5 pixels.
The large improvement in accuracy suggests that better training datasets are needed and that these could improve the accuracy of SPyNet further on general scenes.
 While SPyNet+ft is much more accurate than FlowNet+ft, the latter is fine-tuned on different data.

For the Middlebury \cite{baker2011database} dataset, we evaluate the sequences using the base model SPyNet
as well as SPyNet+ft,  which is fine-tuned on the  Sintel-Final dataset;  the Middlebury dataset itself is too small for fine-tuning. 
SPyNet is significantly more accurate on Middlebury, where FlowNet has trouble with the small motions.
Both learned methods are less accurate than Classic+NL on Middlebury but both are also significantly faster.

\section{Analysis}

\begin{figure}[t]
\centerline{
   \includegraphics[width=\linewidth]{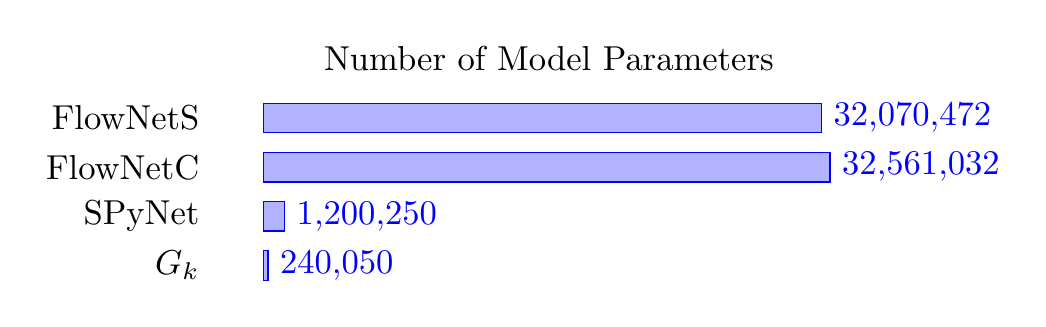}
}
   \caption{Model size of various methods. Our model is 96\% smaller
     than the previous state-of-the-art flow method trained using end-to-end deep learning.}
\label{fig:modelSize}
\end{figure}

\paragraph{Model Size}
Combining spatial pyramids with convnets results in a huge reduction in model complexity.
At each pyramid level, a network, $G_k$, has 240,050 learned parameters.
The total number of parameters learned by the entire network is 1,200,250, with 5 spatial pyramid levels.
In comparison, FlowNetS and FlowNetC \cite{dosovitskiy2015flownet} have 32,070,472 and 32,561,032 parameters respectively. 
SPyNet is about 96 \% smaller than FlowNet (Fig.~\ref{fig:modelSize}).

The spatial pyramid approach enables a significant reduction in model parameters without sacrificing accuracy. 
There are two reasons -- the warping function and learning of residual flow. 
By using the warping function directly, the convnet does not need to
learn it. 
More importantly, the residual learning restricts the range of flow fields in the output space. 
Each network only has to model a smaller range of velocities at each level of the spatial pyramid. 

SPyNet also has a small memory footprint. 
The disk space required to store all the model parameters is 9.7
MB. 
This could simplify deployment on mobile or embedded devices with GPU support.

\begin{figure}[t]
\begin{center}
\subfigure[]{\label{fig:level1layer1}\includegraphics[width=0.485\linewidth]{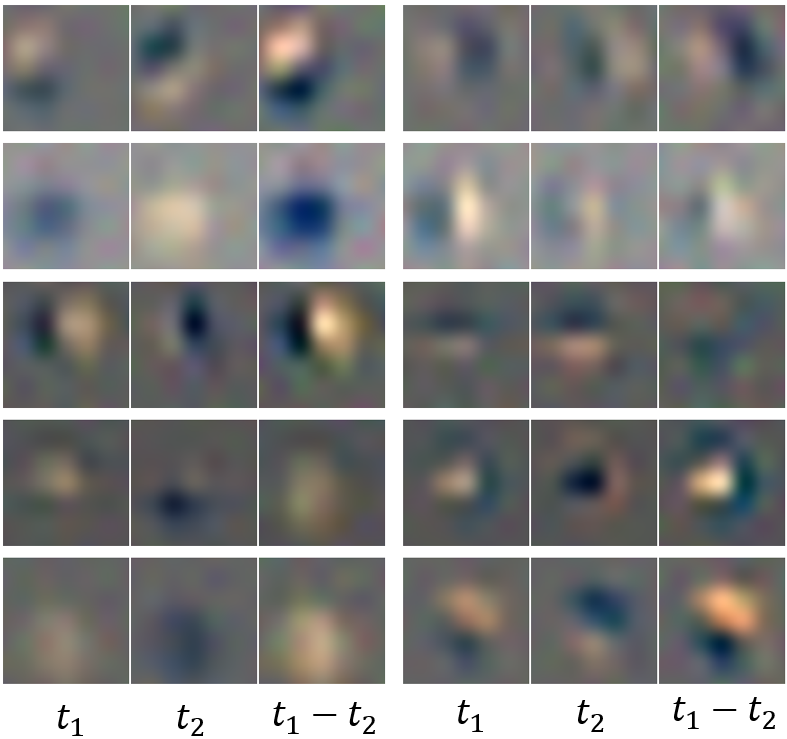}}
~
\subfigure[]{\label{fig:fiterEvolution}\includegraphics[width=0.485\linewidth]{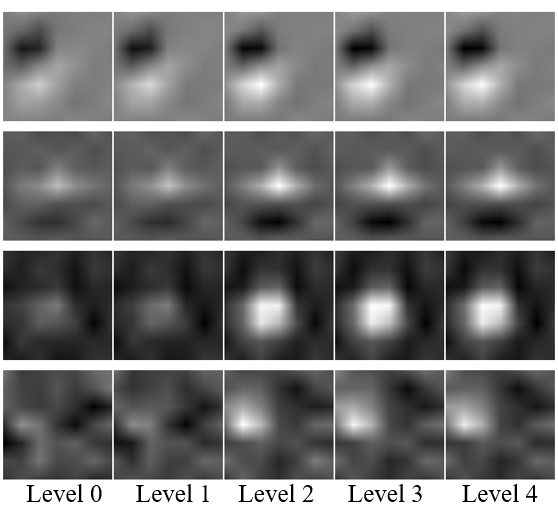}}
\end{center}     
\vspace{-0.1in}
\caption{(a) Visualization of filter weights in the first layer of $G_2$ showing their spatiotemporal nature on RGB image pairs. (b) Evolution of filters across the pyramid levels (from low resolution (0) to high resolution (4))}
\end{figure}
\paragraph{Visualization of Learned Filters.}
 Figure \ref{fig:level1layer1} shows examples of  filters learned by the first layer of the network, $G_2$. 
In each row, the first two columns show the spatial filters that operate on the RGB channels of the two input images respectively. 
The third column is the difference between the two spatial filters hence representing the temporal features learned by our model. 
We observe that most of the spatio-temporal filters in Fig.~\ref{fig:level1layer1} are equally sensitive to all color channels, and hence appear mostly grayscale.
Note that the actual filters are $7\times\/ 7$ pixels and are upsampled for visualization.
 
We observe that many of the spatial filters appear to be similar to traditional Gaussian derivative filters used by classical methods.
These classical filters are hand crafted and typically are applied in the horizontal and vertical direction.
Here, we observe a greater variety of derivative-like filters of varied scales and orientations.
We also observe filters that spatially resemble second derivative or Gabor filters \cite{adelson1985spatiotemporal}.
The temporal filters show a clear derivative-like structure in time.
Note that these filters are very different from those reported in \cite{dosovitskiy2015flownet} (Sup. Mat.), which have a high-frequency structure, unlike classical filters. 

Figure \ref{fig:fiterEvolution} illustrates how filters learned by the network at each level of the pyramid differ from each other. 
Recall that, during training, each network is initialized with the network before it in the pyramid.
The filters, however, do not stay exactly the same with training.
Most of the filters in our network look like rows 1 and 2, where the filters become sharper as we progress towards the finer-resolution levels of the pyramid. 
However, there are some filters that are similar to rows 3 and 4, where these filters become more defined at higher resolution levels of the pyramid. 


\begin{figure}[t]
\centerline{
   \includegraphics[width=0.8\linewidth]{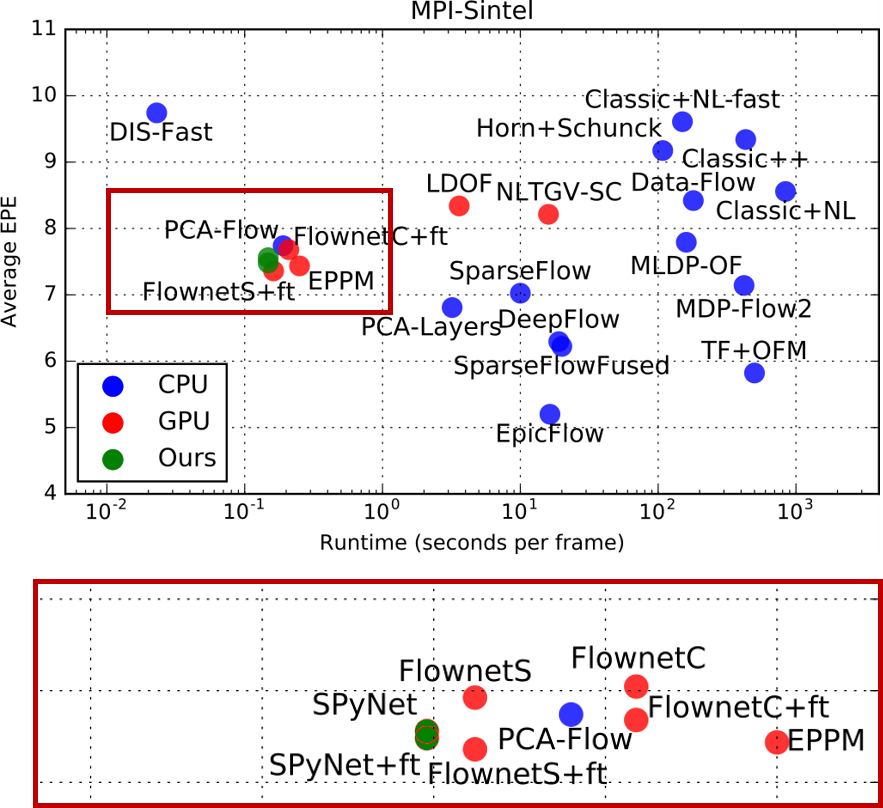}
}
   \caption{Average EPE vs. runtime on MPI-Sintel. 
Zoomed in version on the bottom shows the fastest methods. Times were measured by us. Adapted from \cite{wulff2015efficient}. }
\label{fig:epevtime}
\end{figure}

\paragraph{Speed.}
 Optical flow estimation is traditionally viewed as an optimization problem involving some form of variational inference.
Such algorithms are  computationally expensive, often taking several seconds or minutes per frame.
This has limited the application of optical flow in robotics, embedded systems, and video analysis.

Using a GPU can speed up traditional methods \cite{sundaram2010dense,Werlberger2009:GPUflow} but with reduced accuracy.
Feed forward deep networks \cite{dosovitskiy2015flownet} leverage fast
GPU convolutions and avoid iterative optimization.
Of course for embedded applications, network size is critical (see Fig.~\ref{fig:modelSize}).
Figure \ref{fig:epevtime} shows the speed-accuracy comparisons of several well known methods.
All times shown are measured with the images already loaded in the memory. 
The errors are computed as the average EPE of both the clean and final MPI-Sintel sequences. 
SPyNet offers a good balance between speed and accuracy; no
faster method  is as accurate.

\section{Discussion and Future Work}
Traditional flow methods linearize the brightness constancy equation resulting in an optical flow constraint equation
implemented with spatial and temporal derivative filters.
Sometimes methods adopt a more generic filter constancy assumption \cite{adelson1984pyramid,brox2004high}.
Our filters are somewhat different.  
The filters learned by SPyNet are used in the direct computation of the flow by the feed-forward network.  


SPyNet is small compared with other recent optical flow networks.
Examination of the filters, however, suggests that it might be possible to make it significantly smaller still.
Many of the filters resemble derivative of Gaussian filters or Gabor filters at various scales, orientations, spatial frequencies, and spatial shifts.
Given this, it may be possible to significantly compress the filter bank by using dimensionality reduction or by using a set of analytic spatio-temporal features.
Some of the filters may also be separable.

Early methods for optical flow used analytic spatio-temporal features but, at the time, did not produce good results and the general line of spatio-temporal filtering decayed.
The difference from early work is that our approach suggests the need for a large filter bank of varied filters.
Note also that these approaches considered only the first convolutional layer of filters and did not seek a ``deep'' solution. 
This all suggests the possibility that a deep network of analytic filters could perform well.  
This could vastly reduce the size of the network and the number of parameters that need to be learned.

Note that pyramids have well-known limitations for dealing with large motions \cite{brox2009large,Sevilla:ECCV:2014}.
In particular, small or thin objects that move quickly effectively disappear at coarse pyramid levels, making it impossible to capture their motion.
Recent approaches for dealing with such large motions use sparse matching to augment standard pyramids \cite{brox2009large,weinzaepfel2013deepflow}.
Future work should explore adding long-range matches to SPyNet.
Alternatively Sevilla et al.~\cite{Sevilla:ECCV:2014} define a channel constancy representation that preserves fine structures in a pyramid.
The channels effectively correspond to filters that could be learned.

A spatial pyramid can be thought of as the simple application of a set of linear filters.
Here we take a standard spatial pyramid but one could learn the filters for the pyramid itself.
SPyNet also uses a standard warping function to align images using the flow computed from the previous pyramid level. 
This too could be learned. 



An appealing feature of SPyNet is that it is small enough to fit on a mobile device.  
Future work will explore a mobile implementation and its applications.
Additionally, we will explore extending the method to use more frames (e.g.~3 or 4).  
Multiple frames could enable the network to reason more effectively about occlusion.

Finally, Flying Chairs is not representative of natural scene motions, containing many huge displacements.
We are exploring new training datasets to improve performance on common sequences where the motion is less dramatic.
\section{Conclusions}
In summary, we have described a new optical flow method that combines features of classical optical flow algorithms with deep learning.
In a sense, there are two notions of ``deepness'' here.
First we use a ``deep'' spatial pyramid to deal with large motions.
Second we use deep neural networks at each level of the spatial pyramid and train them to estimate a flow {\em update} at each level.
This approach means that each network has less work to do than a fully generic flow method that has to estimate arbitrarily large motions.
At each pyramid level we assume that the motion is small (on the order of a pixel).
This is borne out by the fact that the network learns spatial and temporal filters that resemble classical derivatives of Gaussians and Gabors.
Because each sub-task is so much simpler, our network needs many fewer parameters than previous methods like FlowNet.
This results in a method with a small memory footprint 
that is faster than existing methods.
At the same time, SPyNet achieves an accuracy comparable to FlowNet, surpassing it in several benchmarks. 
This opens up the promise of optical flow that is both accurate, practical, and widely deployable.

\section*{Acknowledgement}
We thank Jonas Wulff for his insightful discussions about optical flow. 


\end{document}